\documentclass[runningheads]{llncs}

\usepackage{graphicx}
\usepackage{booktabs}
\usepackage{lmodern}
\usepackage{amsmath,amssymb}
\usepackage{microtype}
\usepackage{xcolor}
\usepackage[hidelinks]{hyperref}
\usepackage{underscore} 

\title{When +1\% Is Not Enough: A Paired Bootstrap Protocol for Evaluating Small Improvements}
\titlerunning{Paired Bootstrap Protocol for Small Improvements}
\author{Du Wenzhang}
\institute{Dept. of Computer Engineering, Mahanakorn University of Technology, International\\
College (MUTIC)\\
Bangkok, Thailand\\
\email{dqswordman@gmail.com}}

\begin{document}

\maketitle

\begin{abstract}
Recent machine learning papers often report one--two percentage point improvements from a single run. These gains are measured on standard benchmarks and are highly sensitive to random seeds, data ordering, and implementation details, yet are rarely accompanied by uncertainty estimates or significance tests. It is therefore unclear when a reported ``+1--2\%'' reflects a real algorithmic advance versus noise.

We revisit this problem under realistic compute budgets, where only a few runs are affordable. We propose a simple, PC-friendly protocol based on paired multi-seed runs, bias-corrected and accelerated (BCa) bootstrap confidence intervals, and a sign-flip permutation test on per-seed deltas. The protocol is intentionally conservative and is meant as a guardrail against over-claiming.

We instantiate it on CIFAR-10, CIFAR-10N, and AG News using synthetic ``no-improvement'', small-gain, and medium-gain scenarios. Single runs and unpaired $t$-tests often suggest significant gains for 0.6--2.0 point improvements, especially on text. With only three seeds, our paired protocol never declares significance in these settings. We argue that such conservative evaluation is a safer default for small gains under tight budgets.
\end{abstract}

\section{Introduction}
Single numbers still dominate empirical reporting in machine learning. A typical paper trains a baseline and a new variant once, reports a difference of one or two percentage points on a benchmark, and concludes that the new method ``significantly outperforms'' the baseline. Yet a growing body of work has shown that performance can vary substantially across random seeds, data orderings, and even hardware and library versions~\cite{bouthillier2019unreproducible,bouthillier2021accounting,henderson2018deep,reimers2017reporting,dodge2019show}. Small improvements are particularly fragile: a different seed or slightly different environment may erase or reverse the observed gain.

Despite this, multi-seed reporting and principled uncertainty quantification remain the exception rather than the rule. When authors do repeat experiments, they often apply unpaired $t$-tests that assume independence and approximate normality, conditions that may not hold for small samples and skewed metric distributions~\cite{dietterich1998approximate,demsar2006statistical,dror2018hitchhiker}. Moreover, many practitioners can afford at most a handful of seeds (e.g., three), making classical asymptotic arguments even less convincing.

In this paper we do \emph{not} propose yet another training trick. Instead, we turn the knife toward \textbf{evaluation itself} and ask:

\medskip
\begin{quote}
Given a small number of runs and a small apparent improvement, how can we avoid over-claiming success?
\end{quote}
\medskip

Our answer is a simple protocol that any lab can run on a single consumer GPU. For each of $k$ random seeds, we train both the baseline and the new variant under identical conditions (same data split, same seed, same number of epochs), record the per-seed metric difference, and then analyse these deltas with a BCa bootstrap interval and a sign-flip permutation test. A method is allowed to claim a ``significant improvement'' only if the BCa interval lies entirely above zero and the permutation $p$-value is below a chosen threshold (we use 0.05). Otherwise, the default recommendation is \emph{not} to claim significance.

We study how this protocol behaves in realistic small-gain settings by constructing controlled scenarios on three benchmarks---CIFAR-10, CIFAR-10N, and AG News---with no, small, and medium modifications. We compare three evaluation schemes:
(i) single-run deltas;
(ii) unpaired Welch $t$-tests on independent runs; and
(iii) our paired BCa+permutation procedure.
Empirically we find that single-run comparisons and unpaired tests frequently report apparently significant gains, especially on noisy labels and text~\cite{wei2022learning,zhang2015character}, while our protocol systematically refuses to endorse $0.5$--$2$ percentage-point gains when only three seeds are available.

Our contributions are:
\begin{enumerate}
  \item A simple, paired evaluation protocol for small improvements that combines per-seed deltas, BCa bootstrap confidence intervals, and sign-flip permutation tests into a conservative decision rule.
  \item A controlled case study across three benchmarks and synthetic S0/S1/S2 scenarios, showing how standard practice overstates evidence for small gains, while our protocol behaves as a guardrail against over-claiming.
  \item Practical guidance for authors and reviewers on interpreting small improvements under tight compute budgets.
\end{enumerate}

\section{Related Work}
\subsection{Reproducibility and variance in ML experiments}
Multiple studies have documented substantial run-to-run variance in deep learning experiments, arising from random initialisation, mini-batch sampling, data ordering, and implementation quirks~\cite{bouthillier2019unreproducible,bouthillier2021accounting,henderson2018deep,reimers2017reporting,wei2022learning}. Such variance can reorder model rankings and invalidate conclusions drawn from single runs, especially on modest-size benchmarks.

Reproducibility initiatives and checklists for major conferences encourage authors to report seeds, hyperparameters, and environment details, and to release code when possible~\cite{pineau2021improving,dodge2019show}. These efforts improve transparency but do not by themselves answer the question of how many runs are enough, or how to aggregate a small number of runs into a statistically grounded conclusion. Our work is complementary: we focus not on open-sourcing or environment management, but on the statistical layer on top of a handful of runs.

\subsection{Statistical comparison of learning algorithms}
Classical work on comparing classifiers advocates a variety of tests, including cross-validated $t$-tests, the 5$\times$2CV test, Wilcoxon signed-rank tests, and permutation methods~\cite{dietterich1998approximate,demsar2006statistical,good2005permutation}. These approaches typically assume reasonably large sample sizes (either many datasets or many folds) and i.i.d.\ conditions. In modern deep learning, we often have the opposite: a single dataset, few seeds, and complicated dependence structures.

Bootstrap methods have been proposed as more robust alternatives for non-Gaussian, skewed metrics. The BCa bootstrap adjusts for bias and skewness in the resampled distribution and enjoys good asymptotic properties~\cite{efron1987better,efron1994bootstrap}. Permutation tests provide finite-sample control of type-I error under symmetry assumptions~\cite{good2005permutation}. Our protocol builds on these tools but adapts them to the \textbf{per-seed delta} setting, where we explicitly exploit pairing and accept that the number of deltas is tiny ($k \approx 3$).

\subsection{Multi-seed reporting and paired evaluation}
In NLP and vision, several recent papers have argued for reporting score distributions over random seeds and for stronger statistical testing~\cite{reimers2017reporting,dror2018hitchhiker,dodge2019show}. They show that rankings between models can flip when new seeds are added, and that standard practice often underestimates variance. These works typically use unpaired tests or cross-dataset comparisons. We extend this line by (i) formalising a paired delta + bootstrap + permutation pipeline, and (ii) empirically characterising its behaviour as a conservative filter in realistic small-gain scenarios.

\section{Paired Bootstrap Evaluation Protocol}
\subsection{Problem setting}
We consider two models: a baseline $M_0$ and a variant $M_1$. Given a dataset $D$ and a performance metric $\theta(\cdot)$ (e.g., test accuracy), each training run produces a random outcome due to stochastic optimisation and data sampling. Let $X$ and $Y$ denote the metric for $M_1$ and $M_0$ respectively in one run.

The quantity of interest is the expected improvement
\[
\Delta_{\mathrm{true}} = \mathbb{E}[X - Y].
\]
We aim to answer: Is $\Delta_{\mathrm{true}}$ clearly positive, zero, or negative? Given only a small number of runs $k$, how cautious should we be in claiming $\Delta_{\mathrm{true}} > 0$?

\subsection{Paired multi-seed design}
We adopt a paired design. For each seed $s_i$, $i=1,\dots,k$:
\begin{enumerate}
    \item Fix all randomness (initialisation, data shuffling, augmentation) via seed $s_i$.
    \item Train $M_0$ and evaluate $\theta_{0,i}$ on a fixed test set.
    \item Under the same seed $s_i$, train $M_1$ and evaluate $\theta_{1,i}$.
    \item Record the per-seed delta $\Delta_i = \theta_{1,i} - \theta_{0,i}$.
\end{enumerate}
This mimics a counterfactual experiment: ``holding the world defined by seed $s_i$ fixed, how much better is $M_1$ than $M_0$?''

If $X$ and $Y$ denote the random metrics in a paired run, basic variance algebra gives
\[
\mathrm{Var}(X - Y) = \mathrm{Var}(X) + \mathrm{Var}(Y) - 2\, \mathrm{Cov}(X,Y).
\]
In typical settings, $X$ and $Y$ are positively correlated: both models perform slightly better on easy splits and slightly worse on hard ones, so $\mathrm{Cov}(X,Y) > 0$. Thus pairing reduces the variance of $X-Y$ relative to independent runs. We denote the sample mean of deltas by $\bar{\Delta} = \tfrac{1}{k}\sum_{i=1}^k \Delta_i$.

\subsection{BCa bootstrap confidence intervals}
Given $\{\Delta_i\}$, we estimate uncertainty via the bias-corrected and accelerated (BCa) bootstrap~\cite{efron1987better,efron1994bootstrap}:
\begin{enumerate}
    \item Generate $B$ bootstrap samples by resampling the $k$ deltas with replacement, each time computing the bootstrap mean $\bar{\Delta}^\ast$.
    \item Compute the bias-correction and acceleration terms from the bootstrap and jackknife distributions.
    \item Derive adjusted quantiles for the lower and upper bounds of the $(1-\alpha)$ interval, $[L, U]$.
\end{enumerate}
The BCa interval adjusts for bias and skewness in the empirical distribution of $\bar{\Delta}$ and is particularly useful when $k$ is small and the distribution is asymmetric. Under standard regularity conditions and large $B$, BCa intervals are asymptotically valid.

\subsection{Sign-flip permutation test}
To test $H_0: \Delta_{\mathrm{true}} = 0$, we use a sign-flip permutation test~\cite{good2005permutation}:
\begin{enumerate}
    \item Compute the observed test statistic $T_{\text{obs}} = \bar{\Delta}$.
    \item For each of $P$ permutations, sample random signs $\sigma_i \in \{-1, +1\}$ and compute $T^{(p)} = \tfrac{1}{k}\sum_{i=1}^k \sigma_i \Delta_i$.
    \item Estimate the two-sided p-value as $p = \tfrac{1 + \#\{p : |T^{(p)}| \ge |T_{\text{obs}}|\}}{1 + P}$.
\end{enumerate}
Under symmetry of the $\Delta_i$ distribution, this test controls type-I error exactly at finite $k$. In practice, it behaves conservatively when $k$ is very small, which we accept and even desire for our guardrail purpose.

\subsection{Decision rule}
Our final decision rule is intentionally strict:
\begin{itemize}
    \item Call ``significant improvement'' only if the BCa interval $[L, U]$ satisfies $L > 0$ and the permutation p-value $p < \alpha$ (we use $\alpha = 0.05$).
    \item In all other cases---including ambiguous intervals or borderline p-values---our default recommendation is: \emph{Do not claim a statistically significant improvement.}
\end{itemize}
This rule is designed as a brake, not an accelerator: when evidence is weak or sample size is tiny, it pushes authors toward neutrality.

\section{Experimental Setup}
\subsection{Datasets and models}
We evaluate the protocol on three benchmarks, chosen to cover both vision and text, and both clean and noisy labels.

\paragraph{CIFAR-10.}
50k training and 10k test 32$\times$32 colour images over 10 classes~\cite{krizhevsky2009learning}. We train a standard ResNet-18 with SGD (momentum 0.9, batch size 128) and moderate data augmentation (random crop and horizontal flip).

\paragraph{CIFAR-10N.}
CIFAR-10 with human-annotated noisy labels~\cite{wei2022learning}. We train the same ResNet-18 backbone but use noisy labels for training and the clean test set for evaluation.

\paragraph{AG News.}
A text classification dataset with 120k training and 7.6k test news headlines in four categories~\cite{zhang2015character}. We join title and description, tokenise with a simple word tokenizer, build a 20k-word vocabulary, and train a lightweight TextCNN model, a standard convolutional architecture used in prior work on text classification and vision~\cite{zhang2015character,szegedy2016rethinking}.

All experiments are run on a single consumer GPU (NVIDIA GeForce RTX 5090, CUDA 12.8) with PyTorch 2.8.0 under Windows 10, using Python 3.9.21 and \texttt{num\_workers=0}. For each configuration we use exactly the same training schedule across protocols and seeds.

\subsection{Synthetic scenarios S0/S1/S2}
For each dataset we define scenarios that approximate no, small, and medium gains:

\begin{itemize}
  \item \textbf{S0 (no-improve).} Baseline and variant are identical. This approximates $\Delta_{\mathrm{true}} = 0$ and is used to measure false positive behaviour.
  \item \textbf{S1 (small gain).}
  \begin{itemize}
    \item CIFAR-10 and CIFAR-10N: add label smoothing with $\varepsilon = 0.05$ to the cross-entropy loss.
    \item AG News: TextCNN with label smoothing $\varepsilon = 0.05$.
  \end{itemize}
  \item \textbf{S2 (medium gain).}
  \begin{itemize}
    \item CIFAR-10 and CIFAR-10N: combine label smoothing with stronger augmentation.
    \item AG News: label smoothing plus word dropout (10\% tokens dropped to \textsc{PAD} during training).
  \end{itemize}
\end{itemize}

Table~\ref{tab:scenarios} summarises the datasets and scenarios. All runs use $k=3$ paired seeds.

\begin{table}[t]
  \centering
  \caption{Datasets and synthetic scenarios. Intended $\Delta$ is the rough expected gain in percentage points.}
  \label{tab:scenarios}
  \begin{tabular}{lll p{4.4cm} c}
    \toprule
    Dataset & Modality & Scenario & New vs.\ old configuration & Intended $\Delta$ \\
    \midrule
    CIFAR-10 & vision & S0 (no-improve) &
      Identical ResNet-18 baseline & $\approx 0$ \\
    CIFAR-10 & vision & S1 (small gain) &
      + label smoothing ($\varepsilon = 0.05$) & 1--2 \\
    CIFAR-10 & vision & S2 (medium gain) &
      + strong augmentation + label smoothing & $\approx 1$ \\
    CIFAR-10N & vision/noisy & S1 (small gain) &
      Noisy labels, + label smoothing & $\approx 1$ \\
    CIFAR-10N & vision/noisy & S2 (medium gain) &
      Noisy labels, + strong augmentation + LS & $\approx 1$ \\
    AG News & text & S0 (no-improve) &
      Identical TextCNN baseline & $\approx 0$ \\
    AG News & text & S1 (small gain) &
      + label smoothing ($\varepsilon = 0.05$) & 0.5--1 \\
    AG News & text & S2 (medium gain) &
      + LS + word dropout (0.1) & $\approx 1$ \\
    \bottomrule
  \end{tabular}
\end{table}

\subsection{Evaluation protocols compared}
We compare three evaluation schemes:
\begin{enumerate}
  \item \textbf{Single.} A single baseline run vs.\ a single variant run. We report only the difference in test accuracy, with no uncertainty measure.
  \item \textbf{Unpaired $t$-test ($k=3$).} We train 3 independent runs of the baseline and 3 independent runs of the variant, ignoring pairing, and apply Welch's $t$-test on the two sets of accuracies~\cite{dietterich1998approximate,demsar2006statistical,dror2018hitchhiker}.
  \item \textbf{Paired BCa+perm ($k=3$).} Our proposed protocol: three paired runs, BCa confidence interval on deltas, and sign-flip permutation test, with the strict decision rule in Section~3.
\end{enumerate}

For each protocol we record per-scenario means, intervals, p-values, and decisions. Tables and figures in the paper are generated directly from the logged CSV files, in the spirit of transparent reporting advocated in prior work~\cite{reimers2017reporting,dodge2019show}.

\section{Results}
\subsection{Main quantitative results}
Table~\ref{tab:main-results} summarises the small-gain experiments across datasets and scenarios. Each row corresponds to one dataset--scenario pair and compares:
(i) a single-run delta;
(ii) the mean delta and BCa 95\% interval under our paired protocol; and
(iii) permutation and unpaired $t$-test p-values.

\begin{table*}[t]
  \centering
  \caption{Summary of small-gain experiments (percentage-point test accuracy). 
  $\Delta_{\text{single}}$ is the single-run difference.
  $\Delta_{\text{paired}}$ and the BCa interval come from our paired protocol (3 seeds).
  $p_{\text{perm}}$ is the paired sign-flip permutation p-value and
  $p_{\text{unpaired}}$ is Welch's unpaired $t$-test p-value.}
  \label{tab:main-results}
  \begin{tabular}{lllrrrr}
    \toprule
    Dataset & Scenario & Type &
    $\Delta_{\text{single}}$ &
    $\Delta_{\text{paired}}$ (BCa 95\% CI) &
    $p_{\text{perm}}$ &
    $p_{\text{unpaired}}$ \\
    \midrule
    CIFAR-10   & S0 & no-improve &
      0.00 & 0.00 [0.00, 0.00] & 1.00  & 1.00 \\
    CIFAR-10   & S1 & small gain &
      2.29 & 1.82 [0.37, 2.62] & 0.25  & 0.17 \\
    CIFAR-10   & S2 & medium gain &
      0.32 & 0.93 [0.32, 1.47] & 0.25  & 0.12 \\
    CIFAR-10N  & S1 & small gain &
      1.88 & 1.02 [-1.26, 2.25] & 0.50 & 0.39 \\
    CIFAR-10N  & S2 & medium gain &
      0.66 & 0.93 [0.35, 1.41] & 0.24 & 0.09 \\
    AG News    & S0 & no-improve &
      0.00 & 0.00 [0.00, 0.00] & 1.00  & 1.00 \\
    AG News    & S1 & small gain &
      0.46 & 0.64 [0.46, 0.79] & 0.24  & 0.03 \\
    AG News    & S2 & medium gain &
      0.76 & 1.03 [0.76, 1.20] & 0.24  & 0.003 \\
    \bottomrule
  \end{tabular}
\end{table*}

Several patterns emerge.

First, the S0 rows confirm a sanity check: in all no-improvement scenarios, the average delta is exactly zero, the BCa interval degenerates to $[0, 0]$, and both permutation and unpaired tests give $p = 1.0$. Our protocol does not introduce spurious differences.

Second, for S1 and S2 the single-run deltas frequently look attractive: between $+0.46$ and $+2.29$ percentage points. On AG News S2, for example, a single run shows a +0.76pp gain; on CIFAR-10 S1 some runs exceed +2pp.

Third, unpaired $t$-tests on three seeds tend to be optimistic. On AG News, unpaired p-values are 0.03 (S1) and 0.003 (S2), which many papers would describe as ``significant'' or even ``highly significant''. On CIFAR-10N S2 the unpaired p-value is 0.09, suggestive but not conclusive.

In contrast, our paired BCa+perm protocol is consistently conservative at $k=3$. It never declares significance:
all permutation p-values are either 0.24, 0.25, or larger, even when BCa intervals are strictly positive (e.g., AG News S2 with $[0.76, 1.20]$).
This is a direct consequence of the small discrete sample of deltas and our strict decision rule, and we intentionally embrace this behaviour as a guardrail.

\subsection{Effect sizes and p-values across tasks}

\begin{figure}[t]
  \centering
  \includegraphics[width=\linewidth]{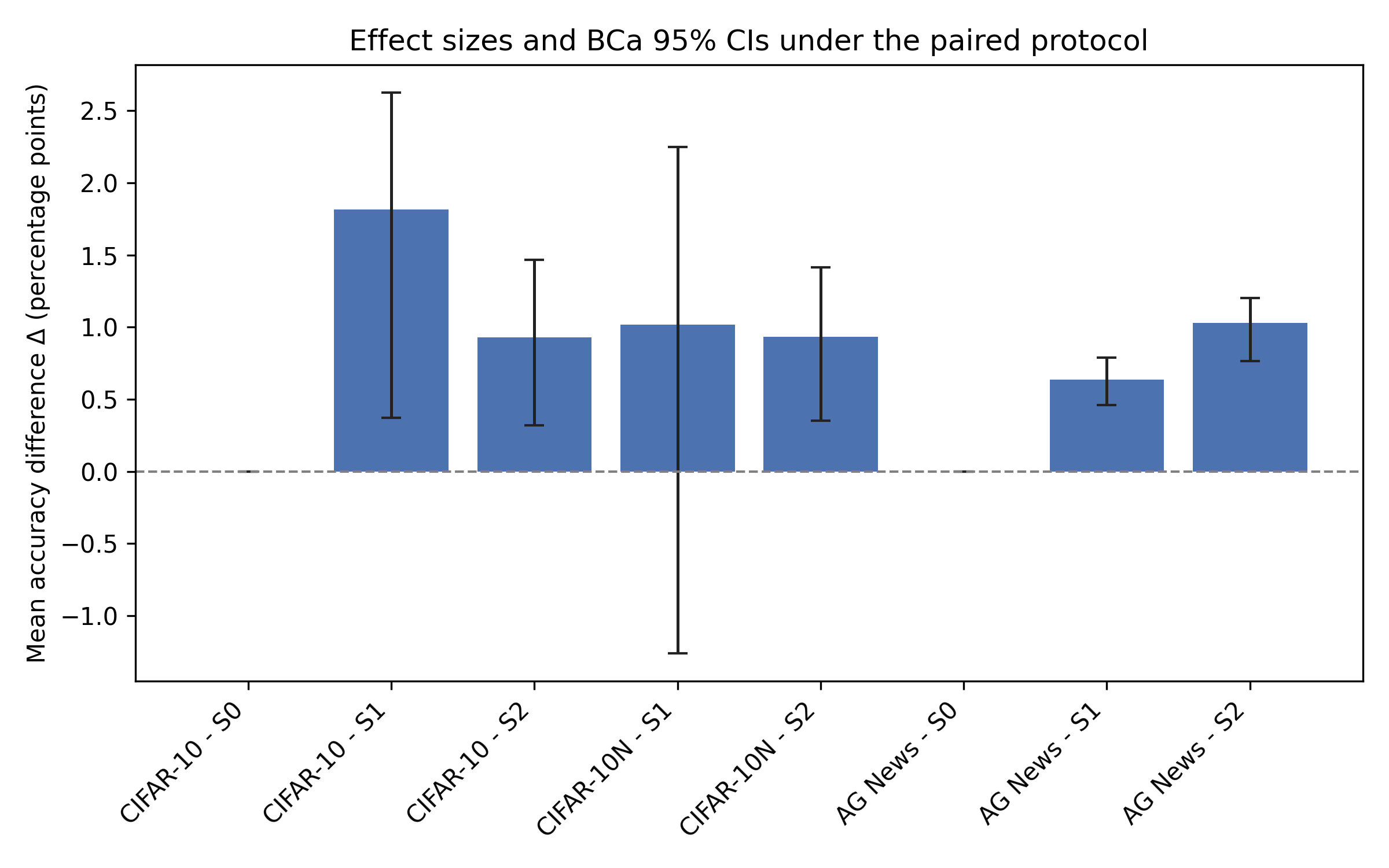}
  \caption{Effect sizes and BCa 95\% confidence intervals under the paired protocol. Bars show $\Delta_{\text{paired}}$; whiskers show BCa intervals; the dashed line marks zero.}
  \label{fig:effect-sizes}
\end{figure}

Figure~\ref{fig:effect-sizes} visualises the mean improvements and BCa 95\% intervals under the paired protocol. S0 bars cluster tightly around zero, confirming no systematic bias. S1/S2 bars lie mostly between +0.5 and +2.0pp, with CIFAR-10N showing the widest intervals due to label noise. Overall effect sizes are small and uncertainty is non-trivial, which supports our view that aggressive significance claims are risky at $k=3$.

\begin{figure}[t]
  \centering
  \includegraphics[width=\linewidth]{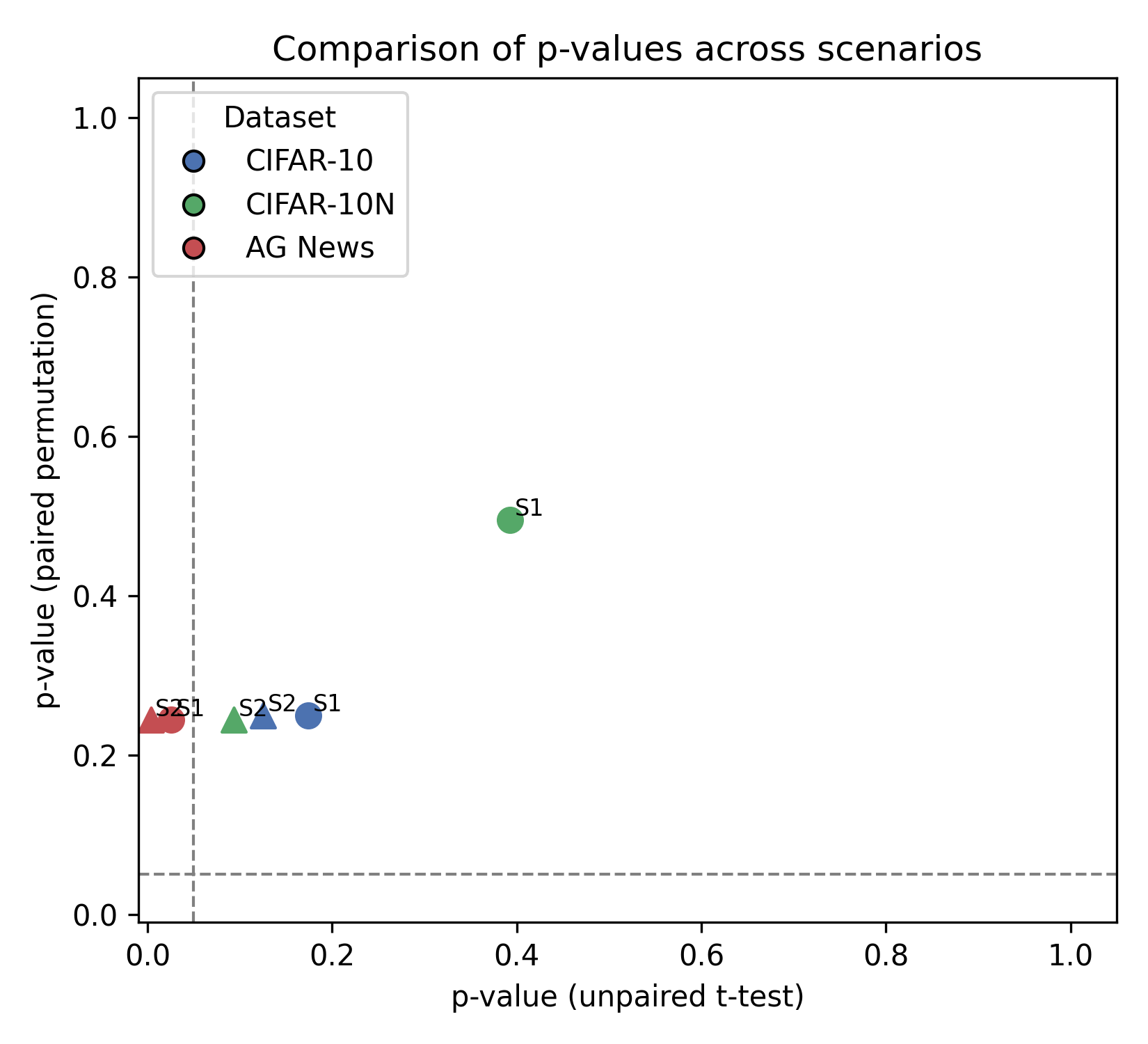}
  \caption{Comparison of p-values across scenarios: unpaired $t$-test (x-axis) vs.\ paired permutation (y-axis). Colours denote datasets; markers denote S1/S2. Dashed lines mark the 0.05 threshold.}
  \label{fig:pvalues}
\end{figure}

Figure~\ref{fig:pvalues} compares unpaired $t$-test p-values (x-axis) with paired permutation p-values (y-axis) across S1/S2 scenarios. Points below the horizontal dashed line and to the left of the vertical line would correspond to cases where both tests call significance. In practice, all S1/S2 points lie in the region where unpaired p-values are small (sometimes $<0.01$) but paired p-values remain around $0.24$. Thus, at $k=3$ our protocol systematically refuses to turn suggestive trends into ``wins''.

\subsection{Case study: CIFAR-10 learning curves}
\begin{figure}[t]
  \centering
  \includegraphics[width=\linewidth]{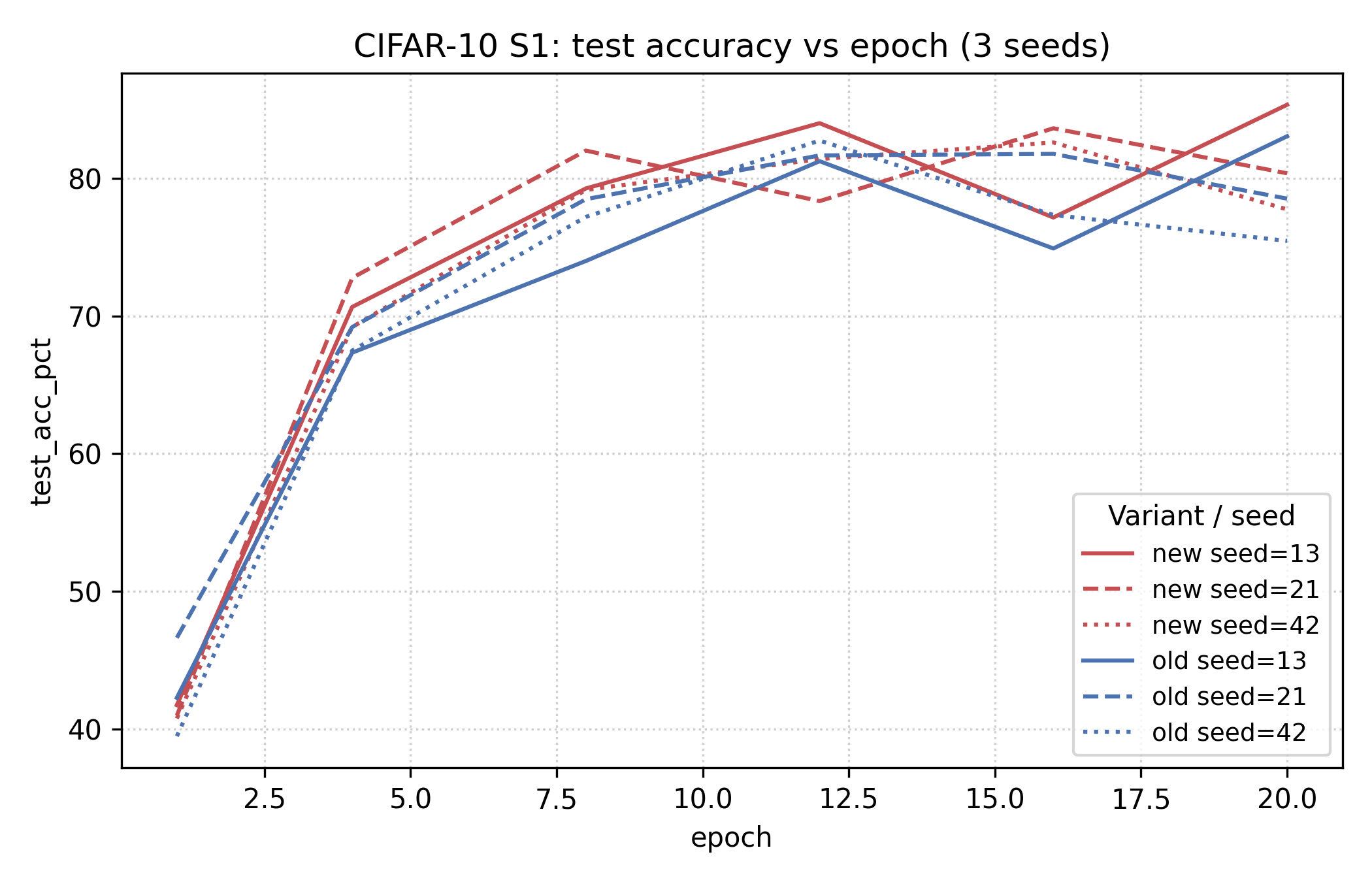}
  \caption{CIFAR-10 S1 learning curves: test accuracy vs.\ epoch for three seeds, baseline (old) vs.\ variant (new).}
  \label{fig:cifar10-s1}
\end{figure}

To illustrate how small gains emerge over training, Figure~\ref{fig:cifar10-s1} plots test accuracy vs.\ epoch for CIFAR-10 S1 (label smoothing), for three seeds and both baseline and variant. The variant consistently tracks slightly above the baseline after the early epochs, with end-of-training gaps between 1--3pp depending on the seed. However, the curves also show non-monotonicity and seed-to-seed variability: some seeds close the gap late, others diverge. Our protocol interprets this as suggestive but not conclusive evidence at $k=3$.

\section{Discussion and Limitations}
Our study argues for a simple but strong practical position: if your improvement is on the order of one to two percentage points and you only have three seeds, you should expect a careful paired bootstrap protocol to say: ``This evidence is too weak to claim significance.''

\paragraph{When is the protocol most useful?}
The protocol is most useful when compute budgets limit you to very few seeds (3--5), effect sizes are small ($\leq 2$pp), and you want a principled way to avoid fooling yourself or your readers. In such settings, our guardrail helps ensure that papers do not over-interpret fragile differences.

\paragraph{Conservatism vs.\ power.}
At $k=3$ the permutation test is deliberately conservative: the discrete null distribution makes it hard to obtain very small p-values. One could relax our decision rule by requiring only CI$>0$ \emph{or} $p<0.05$, which would mark several S1/S2 scenarios as ``significant''. We intentionally adopt the stricter ``CI$>0$ \& $p<0.05$'' rule to prevent over-claiming at minimal $k$ and to encourage richer reporting. When larger budgets allow $k\geq 5$, the same protocol becomes less conservative and can detect moderate effect sizes with reasonable power.

\paragraph{Limitations.}
There are clear limitations. First, we only study small models on CIFAR-10/10N and AG News; larger-scale tasks may exhibit different variance structure. Second, we focus on accuracy; extending the protocol to fairness, robustness or calibration metrics is straightforward in principle but may raise issues with multiple testing and dependency. Third, the permutation test assumes approximate symmetry of the delta distribution; severe skewness may require alternative non-parametric tests or Bayesian treatment. Finally, our protocol does not replace domain judgment: a statistically significant +1\% may or may not be practically important.

\paragraph{Practical recommendations.}
Based on our findings, we recommend: (1) always report deltas across multiple seeds; (2) exploit pairing by training baseline and variant under the same seeds and data splits; (3) use bootstrap intervals and permutation p-values, not only parametric $t$-tests, especially when $k$ is small or metrics are skewed; (4) be explicit about decision rules; and (5) when in doubt, under-claim.

\section{Ethical and Societal Considerations}
Our work concerns evaluation methodology and uses only standard public datasets (CIFAR-10, CIFAR-10N, AG News) that contain no personally identifiable information. It does not involve human subjects or sensitive attributes. The primary societal impact is indirect and positive: if widely adopted, conservative small-gain evaluation could reduce publication of spurious improvements and encourage more rigorous empirical practice. At the same time, stricter evaluation might raise the bar for publishing modest but practically valuable tweaks. Statistical significance should not be the only criterion of value: effect sizes, computational cost, and domain needs should all be considered. Our protocol is meant as a tool to calibrate claims, not as a gatekeeping mechanism.

\section{Reproducibility}
All experiments are designed to be reproducible on a single consumer GPU. We fix random seeds at both the Python and framework level; log system fingerprints (OS, Python, PyTorch, CUDA, GPU model); store per-seed metrics and configuration files; and provide scripts to re-run experiments and regenerate tables and figures. Three key artefacts are:
(i) data loading and training scripts for CIFAR-10/CIFAR-10N (ResNet-18) and AG News (TextCNN);
(ii) an evaluation script that implements the paired BCa+perm protocol and outputs a summary CSV; and
(iii) analysis code that reads the summary CSV and produces Figures~\ref{fig:effect-sizes}--\ref{fig:cifar10-s1} and the main tables.
These artefacts allow other researchers to reproduce our findings, extend the protocol to new tasks, or stress-test it under different sample sizes and decision rules.

\section{Conclusion}
We revisited a mundane but pervasive situation: a new method shows a +1--2 percentage-point improvement on a benchmark under a few random seeds. Should we believe this improvement? Should we call it statistically significant?

Our answer is deliberately cautious. Through a simple paired bootstrap protocol evaluated on CIFAR-10, CIFAR-10N, and AG News, we show that standard practices---single runs and unpaired $t$-tests---readily interpret such gains as significant, especially in noisy or textual settings. In contrast, our protocol acts as a guardrail: it never produces false positives in carefully constructed no-improvement scenarios and systematically declines to endorse small gains when only three seeds are available, even when p-values from unpaired tests look impressive. We view this conservatism not as a weakness, but as a feature: when evidence is weak, the default should be ``no strong conclusion yet.'' We hope this work helps shift the culture around empirical claims in machine learning, especially for small improvements under tight budgets, and encourages the community to invest as much care in evaluation design as in model design.


\end{document}